\documentclass{article}
\usepackage{microtype}
\usepackage{subfigure}
\usepackage{booktabs} 
\usepackage{lineno,hyperref}
\usepackage{xcolor}
\usepackage{url}            
\usepackage{booktabs}       
\usepackage{amsfonts}       
\usepackage{nicefrac}       
\usepackage{microtype}      
\usepackage{multirow}
\usepackage{makecell}
\usepackage{lipsum}
\usepackage{bm}
\usepackage{cuted}
\usepackage{float}
\usepackage{graphicx}
\usepackage{amsmath}
\usepackage{amssymb}
\usepackage[title]{appendix}
\usepackage{mathtools}
\usepackage{algorithm,algorithmic}
\newcommand{\set}[1]{\mathcal{\uppercase{#1}}}

\usepackage[nohyperref,accepted]{icml2019}

\icmltitlerunning{Self-Organization of Action Hierarchy and Compositionality by Reinforcement Learning with Recurrent Neural Networks}

\begin{document}

\twocolumn[
\icmltitle{Self-Organization of Action Hierarchy and Compositionality\\ by Reinforcement Learning with Recurrent Neural Networks}




\begin{icmlauthorlist}
\icmlauthor{Dongqi Han}{cnru}
\icmlauthor{Kenji Doya}{ncu}
\icmlauthor{Jun Tani}{cnru}
\end{icmlauthorlist}

\icmlaffiliation{cnru}{Cognitive Neurorobotics Research Unit, Okinawa Institute of Science and Technology, Okinawa, Japan}

\icmlaffiliation{ncu}{Neural Computation Unit, Okinawa Institute of Science and Technology, Okinawa, Japan}

\icmlcorrespondingauthor{Jun Tani}{jun.tani@oist.jp}


\vskip 0.3in
]



\printAffiliationsAndNotice{} 

\begin{abstract}
Recurrent neural networks (RNNs) for reinforcement learning (RL) have shown distinct advantages, e.g., solving memory-dependent tasks and meta-learning. However, little effort has been spent on improving RNN architectures and on understanding the underlying neural mechanisms for performance gain. In this paper, we propose a novel, multiple-timescale, stochastic RNN for RL. Empirical results show that the network can autonomously learn to abstract sub-goals and can self-develop an action hierarchy using internal dynamics in a challenging continuous control task. Furthermore, we show that the self-developed compositionality of the network enhances faster re-learning when adapting to a new task that is a re-composition of previously learned sub-goals, than when starting from scratch. We also found that improved performance can be achieved when neural activities are subject to stochastic rather than deterministic dynamics. 
\end{abstract}

\section{Introduction}
\label{Chap-intro}

Reinforcement learning (RL) \citep{sutton1998reinforcement} with neural networks as function approximators, i.e., deep RL, has undergone rapid development in recent years. State-of-the-art RL frameworks have shown proficient performance in various kinds of tasks, from game playing \citep{mnih2016asynchronous, silver2016mastering, silver2017mastering} to continuous robot control \citep{lillicrap2015continuous, wang2016sample, haarnoja2018soft}. While most deep RL studies have employed feed-forward neural networks (FNNs) to solve tasks that can be well modeled by Markov Decision Processes (MDP) \citep{bellman1957markovian}, RL with recurrent neural networks (RNNs) has garnered increasing attention \citep{hausknecht2015deep, heess2015memory, shibata2015reinforcement, zhang2016learning, al2017continuous, vezhnevets2017feudal, ha2018recurrent, kapturowski2018recurrent, wang2018prefrontal, jaderberg2019human}.

One benefit of RNNs comes from their capacity to solve a kind of \textit{Partially Observable MDPs} (POMDP) \citep{aastrom1965optimal} in which optimal decision-making requires information derived from historical observations, i.e. memory-dependent tasks. While the curse of dimensionality \citep{friedman1997bias} occurs if these tasks are modelled into MDPs by including all historic information in the current state, a more tractable way of solving memory-dependent tasks is to leverage the contextual capacity of RNNs as function approximators \citep{schmidhuber1991reinforcement}. \citep{wierstra2007solving, utsunomiya2008contextual, heess2015memory} have shown how RNNs enable successful RL in memory-dependent control tasks. Interestingly, even for tasks that are readily solved by MDPs, such as Atari games \citep{mnih2015human}, extraordinary performance can be achieved using relatively simple algorithms with RNNs \citep{kapturowski2018recurrent}.

Furthermore, RNNs advance \textit{meta-learning} in RL,  defined as an effect that an agent requires statistically less time in learning to solve new tasks, compared to previously learned tasks, provided that the two tasks share some common elements \citep{thrun1998learning, wang2018prefrontal, frans2017meta}. \cite{al2017continuous} showed meta-learning by robotic agents in dynamically changing tasks using recurrent policies, while \cite{wang2018prefrontal} argued that the prefrontal cortex, which contains many recurrent connections, plays an important role in meta-learning.

Despite the success of RL with RNN in relatively simple environments, solving more sophisticated tasks often requires cognitive competency in dealing with hierarchical operations, such as for composition/decomposition of an entire task from a sequence of sub-goals \citep{sutton1999between, dietterich2000hierarchical, bacon2017option}. But very few studies have been devoted to developing hierarchical control utilizing RNN architectures. \cite{vezhnevets2017feudal} showed that multiple levels of RNN controllers with different temporal resolutions can achieve dramatic performance on difficult hierarchical RL tasks. However, their method requires one to assume a particular form of transition model for embedding sub-goals.By contrast, our brains are good at self-developing action hierarchies for various tasks and also take advantage of them, which raises a question in our mind: are there any more basic neural mechanisms for discovering an action hierarchy?

Considering this, \cite{yamashita2008emergence} introduced a multiple timescale RNN (MTRNN) containing fast-context and slow-context neurons, which was inspired by the ideas from cognitive science that the multiple timescales are essential to solve complicated cognitive tasks \citep{newell2001time, huys2004multiple, smith2006interacting}. They conducted an experiment in which a humanoid robot learned to generate different motor behavior to operate an object, from supervised samples. Although the explicit hierarchical structure of the task were not given, it was shown that the slow-context neurons autonomously learned to represent abstracted action primitives, such as touching and shaking the object. However, animals usually do not only extract hierarchies from supervised samples, but also can develop functional action primitives through trial-and-error \citep{badre2010frontal, badre2011mechanisms}, which should be modeled by exploration-based RL. Moreover, we wondered how the learned action primitives can enhance learning novel tasks composed of previously learned sub-goals, which was not discussed in \cite{yamashita2008emergence}.

To this end, the current paper proposes a novel multi-timescale RNN architecture and an off-policy actor-critic algorithm for learing with multiple discount factors. We refer to our framework as \textit{Recurrent Multi-timescale Actor-critic with STochastic Experience Replay} (\textbf{ReMASTER}).We also designed a sequential compositional task for testing the performance of the framework. Two essential proposals in this framework are as follows.


The first is to employ a multiple timescale property in neural activation dynamics \citep{newell2001time, huys2004multiple, smith2006interacting, murray2014hierarchy, chaudhuri2015large, runyan2017distinct}, as well as in the discount factors across different levels in an RNN. Although it has been shown that introduction of multiple-timescale neural activation dynamics in RNNs enhances development of hierarchy in supervised learning \citep{yamashita2008emergence}, such a possibility in RL remains to be investigated. In most RL algorithms, the discount factor (for an MDP) is treated as a single hyper-parameter. However, \citep{tanaka2016prediction, enomoto2011dopamine} have shown that dopamine neurons in mammalian brains encode value functions with different region-specific discount factors. In considering motor control, it is intuitive that detailed motor skills are learned with a faster discounting (on the order of seconds), while abstracted actions for long-term plans require longer timescales. In summary, it is expected that more detailed information processing can autonomously develop at lower levels by incorporating the faster timescale constraints imposed on both neural activation dynamics and the reward discounting. Meanwhile, more abstracted action plans can develop at higher levels with slower timescale constraints.

The second proposal is to introduce stochasticity not only in motor outputs, but also in internal neural dynamics at all levels of RNN for generating exploratory behaviors. This is inspired by the fact that cortical neurons, which play a key role in intelligence, have highly stochastic firing behaviors, both for irregular inter-spike intervals and for noisy firing rates \citep{softky1993highly, beck2008probabilistic, beck2012not, hartmann2015s}. \cite{chung2015recurrent} and \cite{fraccaro2016sequential} have shown that various types of stochastic RNN models can learn to extract probabilistic structure hidden in temporal patterns by using variational Bayes approaches in supervised learning. It was also shown that stochastic FNNs facilitate efficient exploration and improved performance in RL \citep{florensa2017stochastic, fortunato2017noisy}. Therefore, we are interested in whether and how stochastic RNNs promote exploration and extraction of task features. 

ReMASTER integrates these two essential ideas (multiple-timescale property and neuronal stochasticity) with an off-policy advantage actor-critic algorithm, in a model-free manner. We considered a kind of sequential, compositional tasks in which an agent learns to accomplish a set of sub-goals in a specific sequence without being given prior knowledge about the sub-goals. The experimental results using ReMASTER showed that compositionality develops autonomously, accompanied by an emergence of hierarchical representation of actions in the network. More specifically, action primitives for achieving task-relevant sub-goals were acquired in the lower level, characterized by faster timescale dynamics, whereas representation of those sub-goals was observed at the higher level characterized by the slower one. As a consequence of such self-developed hierarchical action control, we can ``manipulate'' the agent to consistently pursuing an undesired sub-goal by clamping high-level RNN states, analogous to animal optogenetic experiments \citep{ramirez2013creating, morandell2017role}.

We further examined performance of ReMASTER by considering a multi-phase relearning task wherein an agent is required to adapt consecutively to new tasks that constitute a re-composition of previous-occurred sub-goals. ReMASTER outperformed other alternatives by showing remarkable performance in relearning cases because it was able to take advantage of previously learned representation about the sub-goals in a compositional way, thanks both to multiple timescales and neuronal stochasticity used in the model.

\section{Methods}

\begin{figure}[h]
    \centering
    \includegraphics[width=0.45\textwidth]{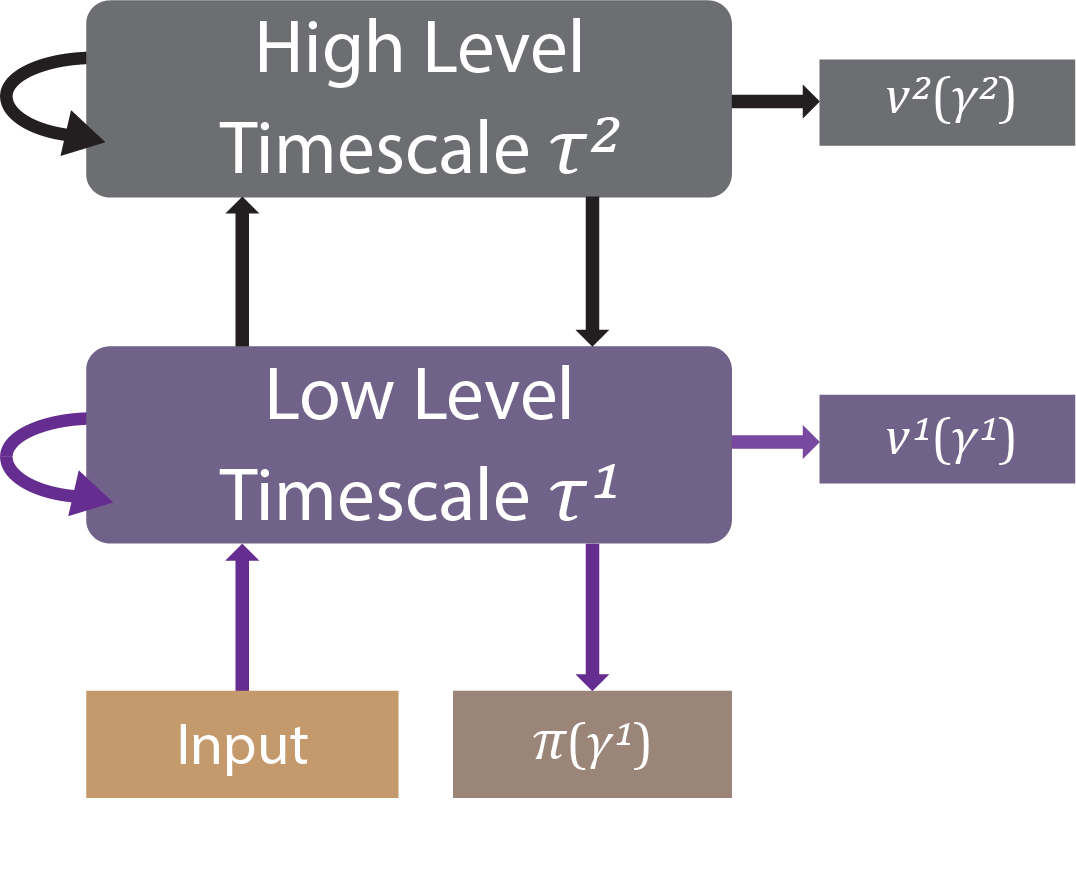}
    \caption{The basic structure of MTSRNN is shown for the case of a 2-level configuration, used in this work. However, additional levels can readily be stacked onto it.}
    \label{fig:MTSRNN-rough}
\end{figure}

We used a multi-level stochastic RNN with level-specific timescales, as a basic network architecture for implementing ReMASTER. This architecture is referred to as \textit{Multiple Timescale Stochastic Recurrent Neural Network} (MTSRNN). Fig.~\ref{fig:MTSRNN-rough} shows the case of a 2-level MTSRNN where and $\gamma^l$ represents the characteristic discount factor at $l$-th level. $v^l$ as the value function at $l$-th level is estimated by temporal difference learning (TD-learning) \citep{sutton1998reinforcement} using the corresponding $\gamma^l$. The policy function with discount factor $\gamma^1$, indicated by $\pi$, is estimated by the lowest level. Also, only the lowest level receives inputs. Note that although the network has multiple timescales of discounting, policy is improved to maximize expected return w.r.t. the lowest discounter factor $\gamma^1$. Learning the higher-level value function(s) $v^{l>1}$ serves as an auxiliary objective, which, nonetheless, we found critical in our experiments.

\subsection{Multiple Timescale Stochastic RNN}
\label{chap:mtsrnn}
Here we describe detailed mechanisms of $L$-levels MTSRNN. We use a super-script $l \in \{1, 2, \dots, L\}$ to indicate the $l$th level, where a smaller $l$ indicates a lower level. Let $\bm u$ and $\bm c$ denote the \textit{hidden states} and the \textit{RNN outputs}, respectively\footnote{We collectively refer to $(\bm u, \bm{c})$ as \textit{RNN states}}, we have
\begin{align}
    \bm{u}^l(t) = & (1-\frac{1}{\tau^l})\bm{u}^l(t-1) + \frac{1}{\tau^l}\left[W^{l-1,l}_{cu} \bm{c}^{l-1}(t) + \nonumber \right. \\ & \left. W^{l,l}_{cu} \bm{c}^{l}(t-1) + W^{l+1,l}_{cu} \bm{c}^{l+1}(t-1)  + \bm{b}_u^l  \right] , \\
    \bm{c}^l(t) = & \tanh \left( \bm{u}^l(t) + \epsilon^l  \bm{\sigma}^l(t) \right),
    \label{eq:c}
\end{align}
where $\bm{c}^{l-1}(t)=\bm{s}(t)$ when $l=1$ is the current sensory input (state) and $\bm{c}^{l+1}$ does not exist for $l=L$. The scale of neuronal noise, $\sigma^l$, can be either a hyper-parameter or adaptive, and $\bm\epsilon^l(t)$ is a diagonal-covariance unit-Gaussian noise, which leads to a stochastic variable $\bm{c}^l(t)$ using the reparameterization trick \citep{kingma2013auto}. The hyper-parameter $\tau^l$ is known as \textit{timescale} of the $l$th level, which determines how fast hidden states vary, for which we usually have $\tau^l < \tau^{l+1}$. Synaptic weights and biases, denoted by $W$ and $\bm{b}$, respectively, are trainable parameters of the neural network.

Value functions can be estimated via a linear connection from each level of the MTSRNN:
$
    v^l(t) = (\bm{w}^l_{cv})^{T} \bm{c}^l(t) + b^l_{cv} .
    \label{eq:value}
$
We focus on continuous action space, so the policy function can be expressed as diagonal Gaussian distributions
$
    \bm{\pi}(t) \sim \set{N}\left(\bm{p}(t), \bm{e}(t)\right),
    \label{eq:pi}
$
where
$
    \bm{p}(t) = \tanh\left(W_{ca} \bm{c}^1(t) + \bm{b}_{ca} \right)
    \label{eq:action}
$
is the expected action assuming that the range of possible actions is $[-1, 1]$, and that  $\bm{e}(t)$ is the exploration scale. 
$
    \bm{e}(t) = \exp{\left[ \frac12 \left(W^l_{ce} \bm{c}^1(t) + \bm{b}^l_{ce} \right)\right]}.
    \label{eq:motor-noise}
$

\subsection{Off-Policy Advantage Actor-Critic}
Recent deep RL studies using actor-critic algorithms with experience replay have achieved remarkable performance in many RL environments \citep{lillicrap2015continuous, wang2016sample, haarnoja2018soft} by learning repetitively from previous experience. Therefore, for better sample efficiency, we use an off-policy actor-critic algorithm \citep{degris2012off}, which can deal with both continuous and discrete action spaces, although we consider continuous control in this work.

Suppose that in each episode, the agent is continuously interacting with the environment. At every step $t$, it experiences a state transition, which can be described by a tuple $(\bm s_t, \bm a_t, \bm s_{t+1}, r_t, done_t, \pi_t)$, where $\bm s$, $\bm a$, $r$, $\pi$ are state (observation), action, reward and policy function, respectively; and the Boolean $done_t$ indicates whether the episode ends at step $t+1$. The agent stores the state transition in a replay buffer. In practice, RNNs require initial states for computing succeeding time development of RNN states. We set initial RNN states to zero at the beginning of each episode. For easier access to initial states during experience replay using RNN, we also recorded $\bm{c}^l(t)$ and $\bm{u}^l(t)$ of the MTSRNN at each step and used them in experience replay\footnote{The recorded RNN states can be incompatible with the current RNN as learning goes on. However, this issue does not impact learning performance much because very old samples are discarded due to limited memory. We took this approach because of its simplicity (More discussion in Appendix~\ref{appendix:initial}).}.

\begin{algorithm*}[tb]
    \caption{\textbf{ReMASTER}}
    \label{algo:overall}
\begin{algorithmic}
    \STATE Initialize the MTSRNN $\set{R}$ and the replay buffer $\set{B}$, global step $t  \leftarrow 0$
    \REPEAT
        \STATE Reset an episode, assign $\set{R}$ with zero initial RNN states
        \WHILE{episode not terminated}
            \STATE Compute 1-step forward of $\set{R}$ to obtain $(\bm{u}^l_t$, $\bm{c}^l_t)$
            \STATE Sample an action $\bm a_t$ from policy $\pi_t(\bm{a} | \bm{c}^1_{t})$ and execute $\bm a_t$
            \STATE Obtain $\bm s_{t+1}$, $r_t$ and $done_t$ from the environment
            \STATE Record ($\bm s_t, \bm a_t, \bm s_{t+1}, r_t, done_t)$, $\pi_t=\pi(\bm{a}_t | \bm{c}^1_{t})$ and $(\bm{u}^l_t$, $\bm{c}^l_t)$ into $\set{B}$
            \IF{$ mod(t, train\_interval) == 0$}
                \STATE Sample sequential training samples
                to update $\set{R}$ by Eq.~\ref{eq:value_update} and \ref{eq:pi_update} 
            \ENDIF
            \STATE $t \leftarrow t+1$
        \ENDWHILE
    \UNTIL{training stopped}
\end{algorithmic}
\end{algorithm*}

For estimation of the critic, we used an off-policy version of the temporal difference (TD) learning algorithm to train value functions of all levels \citep{sutton1998reinforcement, degris2012off}. Knowing that (i) each level has a characteristic timescale $\tau^l$; (ii) $1/(1-\gamma)$ indicates the eigen-timescale of discounting \citep{doya2000reinforcement}, it is natural to set the values of discount factors as
\begin{equation}
    \gamma^l = 1 - \frac{K}{\tau^l},
\end{equation}
where K is a constant to which we assigned a value of 0.16 throughout this work.

Let $\bm\theta$ denote the synaptic weights of the network. At each update, we randomly sample $N$ state transition tuples from memory, and then conduct gradient descent for value functions $v^l$ with learning rate $\alpha_v$,
\begin{equation}
    \bm\theta \leftarrow \bm\theta + \alpha_v \frac{1}{L} \frac{1}{N} \sum_l^L \sum_{i}^N \left[\rho_i \delta_i^l \nabla_{\bm\theta} v^l(\bm{s}_{0:t_i};\bm\theta) \right],  \label{eq:value_update}
\end{equation}
where we have
\begin{equation}
    \rho_i = \frac{\pi(\bm{a}_{t_i}|\bm{s}_{0:t_i};\bm\theta)}{\pi_{t_i}}
\end{equation}
indicating the importance sampling ratio of the $i$th sample, where $\pi_{t_i}$ is the behavior policy obtained from the replay buffer; and
\begin{equation}
    \delta_i^l = r_{t_i} + \gamma^l  v^l(\bm{s}_{0:t_i + 1};\bm\theta) - v^l(\bm{s}_{0:t_i};\bm\theta)
\end{equation}
is the TD-error for the $i$th sample and the $l$th level. Note that the value function $v^l$ and the policy function $\pi$ depend on $\bm{s}_{0:t_i}$ so that backpropagation through time (BPTT) is performed to calculate the gradients. They can also be written as $v^l(\bm c^l_{t_i}; \bm\theta)$ and $\pi(\bm a | \bm c^1_{t_i}; \bm\theta)$, respectively, if $\bm c^l_{t_i}$ has been computed.

To update the policy function, an advantage policy gradient algorithm was used in an off-policy manner \citep{degris2012off}, where the advantage was estimated by 1-step TD error with discount factor $\gamma^1$.
\begin{equation}
    \bm\theta \leftarrow \bm\theta + \alpha_a \frac{1}{N} \sum_{i}^N  \left[\rho_i \delta_i^1 \nabla_{\bm\theta} \log{\bm{\pi}(\bm{a}_{t_i}|\bm{s}_{0:t_i};\bm\theta)} \right], \label{eq:pi_update}
\end{equation}
where $\alpha_a$ is the learning rate for the actor. Algorithm~\ref{algo:overall} shows the overall procedure of ReMASTER. We followed algorithm~\ref{algo:overall} for all experiments in this work.


\section{Results}
We applied 2-levels ReMASTER to a so-called \textit{sequential target-reaching task}. The following explains the details of task designs and simulation results and their analysis for a basic task, followed by those for an extended task that deals with relearning of consecutive task wherein goal changes. For all the experiments performed in this study, if not specified, we used $\tau^1=2$, $\tau^2=8$, $\gamma^1=0.92$, and $\gamma^2=0.98$. The MTSRNN had 100 and 50 neurons in the lower and the higher level, respectively. For exploration, we applied diagonal Gaussian noise to both hidden states and motor actions, and annealed them exponentially w.r.t. episodes. The learning rate was 0.0003 for the critic and 0.0001 for the actor, each using an RMSProp optimizer with decay=0.99. The network was trained once every 2 steps. More implementation details can be found in Appendix~\ref{appendix:detail}.

\subsection{Sequential Target-Reaching Task}

\begin{figure*}
    \centering
    \includegraphics[width=1.0\textwidth]{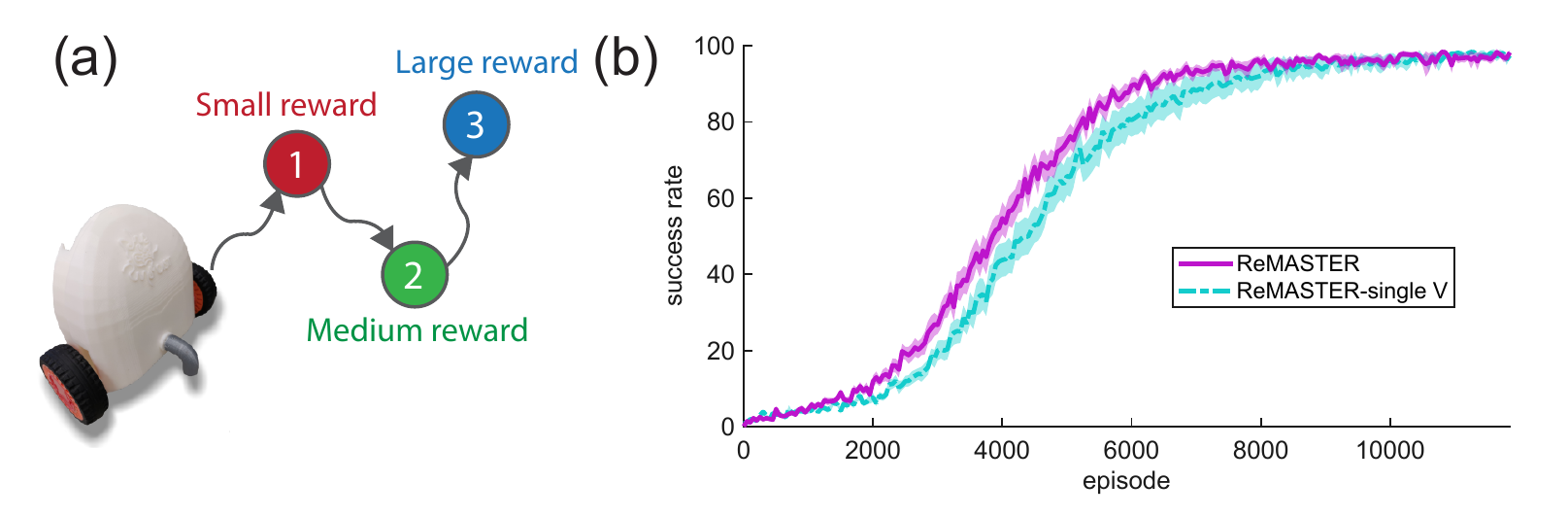}
    \caption{The sequential target-reaching task: (\textbf{a}) Illustration of the task. Configuration of the robot and the targets was randomly initialized in each episode. (\textbf{b}) Performance curve indicated by success rate. ReMASTER-single V is the case in which the higher-level value function was not learned. Data are Mean $\pm$ S.E.M., obtained from 20 repeats.}
    \label{fig:Phase1}
\end{figure*}

We propose a sensory-motor task, inspired by \citep{utsunomiya2008contextual}, referred to as a ``sequential target-reaching task''. As illustrated in Fig.~\ref{fig:Phase1}(a), a two-wheeled robot agent on a 2-dimensional field is required to reach three target positions, in a sequence, red-green-blue, without receiving any signal indicating which target to reach. The action is given by the rotation of its left and right wheels. At the beginning of each episode, the robot and targets are randomly placed on the field. The robot has sensors detecting distances and angles from the three targets, as well as the current-step reward. When it reaches a target in the correct sequence, it receives a one-step reward. The reward is given only if the agent followed the proper sequence, and is given only once for each target. An episode terminates if the agent completes the task or a maximum of 128 steps are taken. To successfully solve the task, the agent needs to develop the cognitive capability to remember ``which target has been reached'', as well as to recognize the correct sub-goal (which can be considered as approaching a target in this task). More details of the task can be found in (Appendix~\ref{appendix:sequential}).

The sequential target-reaching task is of particular interest because it abstracts many real-world tasks in complicated environments, which involve decomposition of a whole task into sub-tasks and execution of each sub-task in a specific sequence. One example is dialing on a classic telephone, where one needs to sequentially choose each number and perform detailed hand movements to dial the number. Mastering this kind of task naturally requires development of hierarchical control of actions. Lower levels acquire skills for action primitives, while higher levels learn to dispatch those action primitives in a specified sequence.

We examined ReMASTER in the sequential target-reaching task. Fig.~\ref{fig:Phase1}(b) shows that ReMASTER can successfully solve this task through self-exploration, achieving more than $95\%$ success rate\footnote{An episode was considered successful when the agent completed the task within 50 steps.} on average after training. We also tested the case in which the higher-level value function $v^2$ is not trained. (ReMASTER-single V in Fig.~\ref{fig:Phase1}(b)), which achieved similar success rate in the end but the learning is relatively slower.

\begin{figure*}
    \centering
    \includegraphics[width=1.0\textwidth]{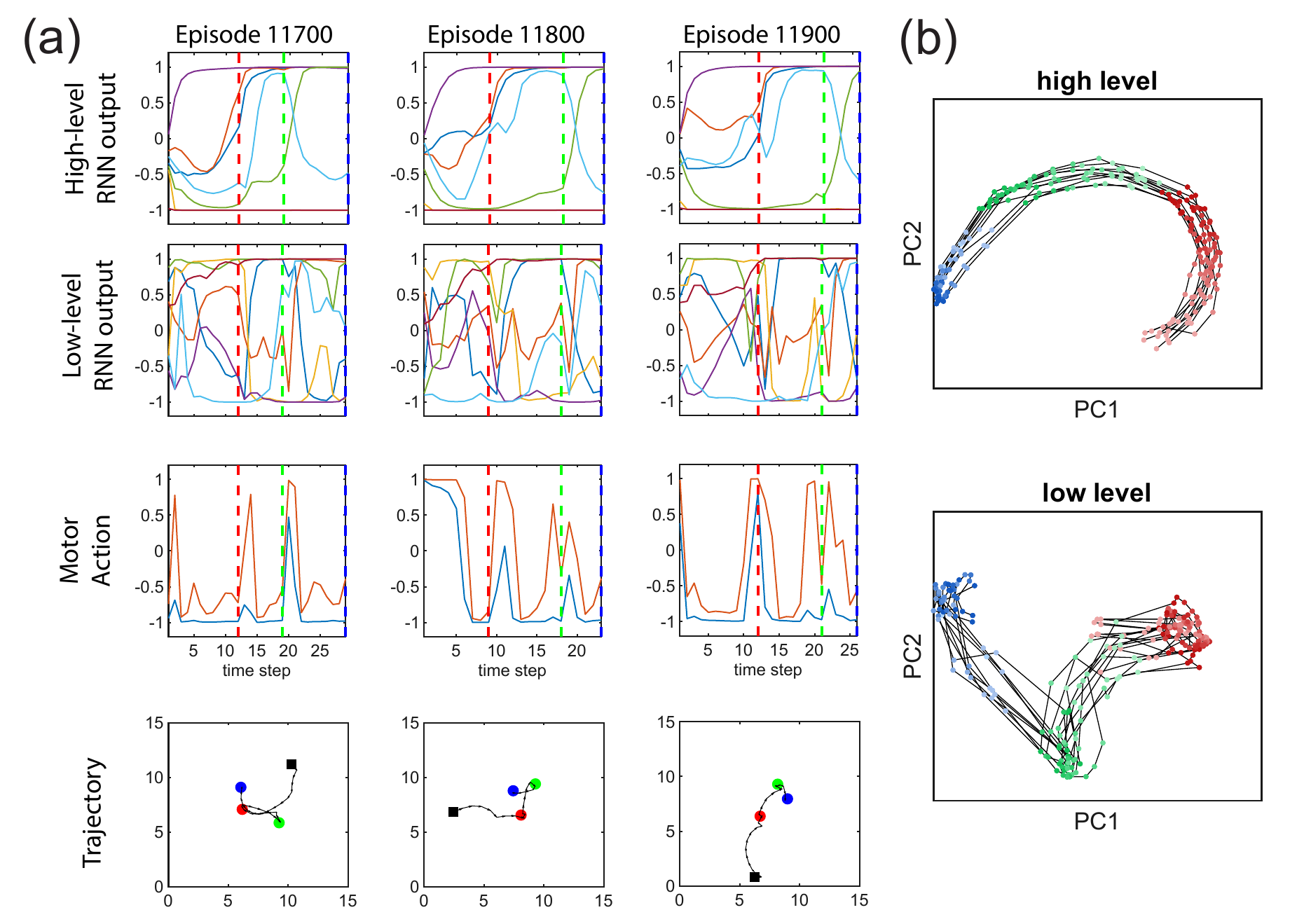}
    \caption{Analysis on the sequential target-reaching task using ReMASTER. (\textbf{a}) Three example episodes showing the behavior of a well-trained ReMASTER agent. The first and second rows show RNN output $\bm{c}^l_t$ of two levels, where the vertical, dashed lines indicate the agent's reaching a target. For clarity, we plotted only ${c}^l_t$ of the first 7 neurons for both levels, with different colors indicating different neurons. The motor actions indicted by velocities of the two wheels are plotted in the third row. The fourth row is the robot's trajectories, where black squares indicate its starting positions and circles are target positions. (\textbf{b}) PCA for visualizing temporal profiles of $\bm{c}^l_t$, using data of the same agent in (a) in episodes 11000, 11100, ..., 11900 (after convergence). Colors mean the agent is approaching to the corresponding targets, where a deeper color means the agent is more closed to the target. Samples from the same episode are linked with black lines.}
    \label{fig:Phase1_ana}
\end{figure*}

However, our major aim was to examine what sorts of internal representation the MTSRNN had developed for achieving sequential hierarchical control after abundant training using ReMASTER. Fig.~\ref{fig:Phase1_ana}(a) shows three examples of how an agent behaved after learning, where the three columns present the behavior of the same agent, but in different episodes. Interestingly, although target configurations and the motor actions (the last 2 rows of Fig.~\ref{fig:Phase1_ana}(b)) were completely different in these episodes, high-level neurons showed relatively similar temporal profiles of RNN outputs $\bm{c}^l_t$, as plotted in the first row of Fig.~\ref{fig:Phase1_ana}(a). In contrast, this feature was less obvious in the lower level (second row of Fig.~\ref{fig:Phase1_ana}(a)). Although we only show one agent here, this result is statistically significant (Section~\ref{consistency}). This result suggests that an MTSRNN with slower dynamics in the higher level enhanced development of a consistent representation, accounting for a given sub-goal structure through abstraction in the higher level, whereas the lower level dealt with details of motor control depending on object configuration in the field in each episode. Consistency in representing sub-goals of the higher level can also be demonstrated by conducting PCA on RNN outputs of the two levels of the MTSRNN after convergence (Fig.~\ref{fig:Phase1_ana}(b)). We can see that the high-level RNN outputs showed a consistent, sequence-like representation of sub-goals accounted by its slower dynamics, whereas the lower level showed a more divergent representation since it needs to generate each different maneuvering trajectory. Hence, we saw an emergence of action hierarchy using ReMASTER.

\subsection{Consecutive Relearning Task}
\label{chap:phase23result}
Since our previous analysis indicated that the low level learns action primitives for achieving each sub-goal, relearning to solve a new task that is a re-composition of previously learned sub-goals in a different sequence, should be much more efficient than starting from scratch.

\begin{figure*}
    \centering
    \includegraphics[width=0.9\textwidth]{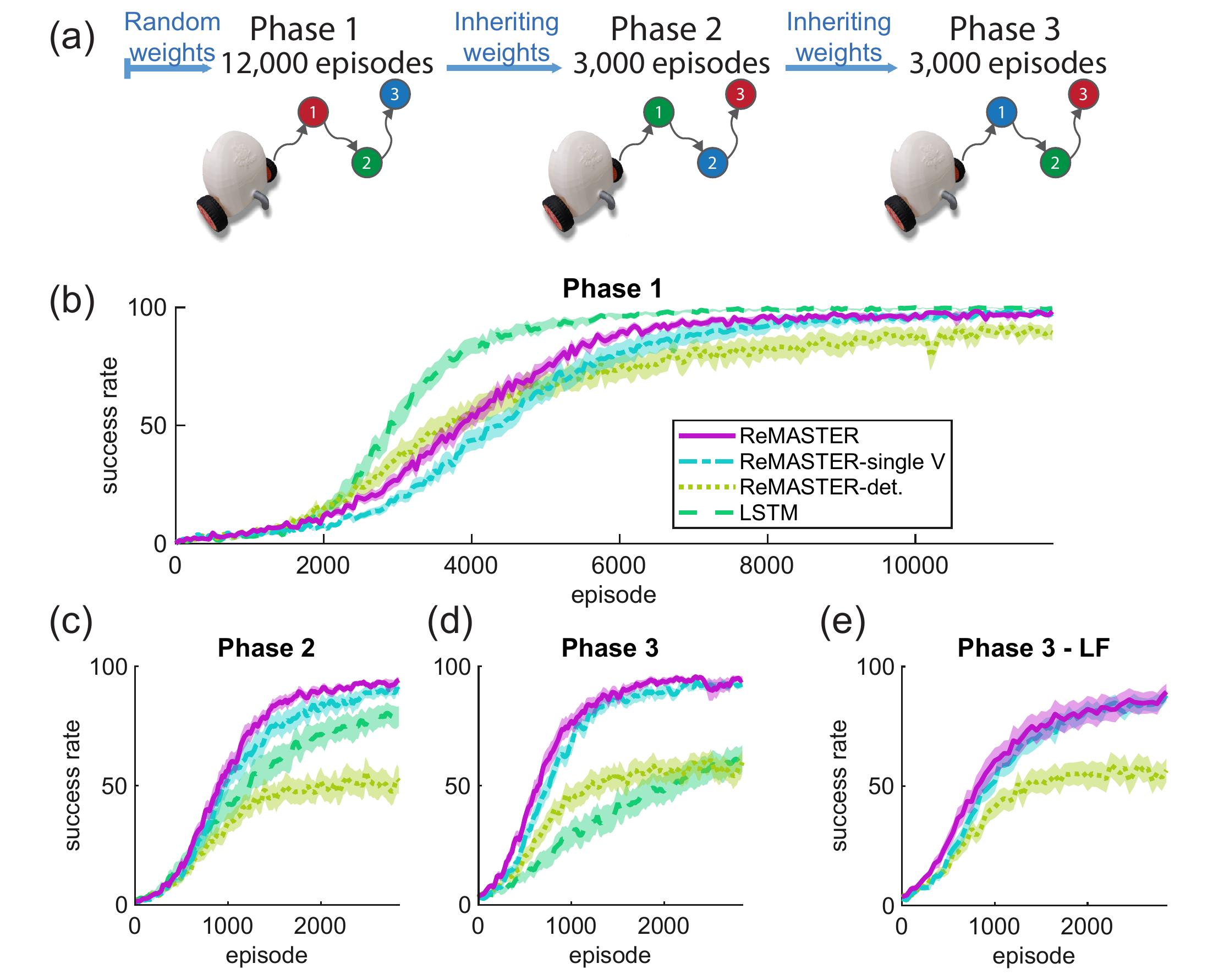}
    \caption{The consecutive relearning task: (\textbf{a}) Illustration of the task. (\textbf{b-d}) Performance curves for all phases, plotted in the same way as Fig.~\ref{fig:Phase1}(b). ReMASTER-det. stands for the case in which all the neurons followed deterministic dynamics, and LSTM is the alternative using the same algorithm but the network was a single-layer LSTM. \textbf{(e)} Performance curve of phase 3 with the lower-level synaptic weights frozen (Phase 3 - LF).}
    \label{fig:Phase2}
\end{figure*}

By considering this, we carried out another experiment, in an extended version of the sequential target-reaching task, referred to as an ``consecutive relearning task'' (Fig.~\ref{fig:Phase2}(a)). In this task, the robot agent was required to adapt consecutively to changed task goals (or more specifically, changed reward functions and termination conditions) by relearning. The consecutive relearning task consisted of 3 different phases. Phase 1 corresponded to the original red-green-blue sequential target-reaching task. Phases 2 and 3 appeared as novel re-compositions of sub-goals, where the required sequences are green-blue-red and blue-green-red, respectively. While  phase 1 had 12,000 episodes, there were only 3,000 episodes in phase 2 or 3. 

We performed experiments on this task in a lifelong learning manner \citep{thrun1994learning, silver2013lifelong}. We maintained the same learning algorithm and hyper-parameters throughout all 3 phases. Synaptic weights were continuously updated without resetting throughout the experiment. At the beginning of each phase, motor and neuronal noise scale were reset and the replay buffer was cleared. We additionally compared performance using ReMASTER to two alternatives, in order to examine the importance of neuronal stochasticity and intrinsic timescale hierarchy. One alternative is the deterministic version of ReMASTER in which there was only motor noise for exploration, but no noise was applied to neurons (ReMASTER-det. in Fig.~\ref{fig:Phase2}(b-d)). Another alternative used the same algorithm, but replaced the MTSRNN with a single-layer LSTM (LSTM  in Fig.~\ref{fig:Phase2}(b-e)) using $\gamma=1-\sqrt{(1-\gamma^1)(1-\gamma^2)}=0.96$, but we got similar performance for $\gamma=\gamma^1$ or $\gamma^2$). The LSTM network contained 75 cells so that the number of parameters is similar to that of the MTSRNN.


Results are illustrated in Fig.~\ref{fig:Phase2}(b-d), which shows task performance in terms of success rate in three different phases. Several conclusions can be drawn from these results\footnote{Although we used tuned hyperparameters for better performance, these conclusions indeed hold for different choice of hyperparameters (Appendix.~\ref{appendix:rp})}.

First, for ReMASTER and ReMASTER-single V, the relearning cases of phases 2, 3 (Fig.~\ref{fig:Phase2}(c,d)) starting with previously trained synaptic weights achieved much better sample efficiency than the case of phase 1  which was done from scratch (Note that there were only 3,000 episodes in phases 2,3, whereas there were 12,000 episodes in phase 1. Rigorous comparison is left to Appendix~\ref{appendix:control}.). We consider that this resulted from compositionality during action hierarchy development, which enabled a flexible re-composition of sub-goals, so that the agents could rapidly adapt to relearning tasks. 

Second, ReMASTER significantly and consistently outperformed ReMASTER-det. in all the three phases (Fig.~\ref{fig:Phase2}(b-d)). One possible reason is that stochastic neurons could prevent the network from over-fitting, thereby enhancing network flexibility. Another is that neuronal noise can lead to larger exploration in the hidden state space \citep{shibata2015reinforcement, fortunato2017noisy}, which results in a greater likelihood of finding adequate neural representation in the higher level, which fits with newly appeared re-composition tasks.

Third, ReMASTER also addressed consistent performance advantage over ReMASTER-single V. (Fig.~\ref{fig:Phase2}(b-d)). Recall that policy is learned to optimize the expected return with discount factor $\gamma^1$. Our results suggested it could be beneficial to learn value functions with multiple discounting, which agrees with the findings that mammalian brains are doing the same thing \citep{tanaka2016prediction, enomoto2011dopamine}.

Finally, ReMASTER and ReMASTER-single V showed a performance advantage over the LSTM alternative in phases 2, 3
, although LSTM achieved great performance in phase 1 (Fig.~\ref{fig:Phase2}(b-d)). We consider the performance degradation of LSTM in phases 2,3 is because of the mixed representation of sub-goal sequencing and detailed motor skills in one level. This created difficulty in relearning sub-goal sequencing while reusing low-level skills. 
In contrast, ReMASTER provided flexible compositionality that enables these two levels of control to be better segregated in different levels in MTSRNN. Although biological plausibility of our approach is arguable, this result may underlie a potential reason of why we have many separated, timescale-distinct brain regions working for multiple levels of functions \citep{murray2014hierarchy, runyan2017distinct, wang2018prefrontal}.

\subsection{Learning to Solve New Tasks with Low-Level Weights Frozen}
The previous results (Fig.~\ref{fig:Phase2}(b-d)) were obtained when both the higher and the lower level synaptic weights were continually trained throughout the task. However, if the lower level had acquired necessary motor skills for achieving the sub-goals, it should be possible for the agent to learn to solve new tasks by updating only the higher level. 

Therefore, we conducted another simulation on the consecutive relearning task using ReMASTER, in which low-level synaptic weights (purple connections in Fig.~\ref{fig:MTSRNN-rough}) were frozen in phase 3, as inherited at the end of phase 2. The ReMASTER and ReMASTER-single V agents showed remarkable learning effectiveness in phase 3 (Fig.~\ref{fig:Phase2}(e)), whereas the ReMASTER-det. agents could improve their policy but the learning was less efficient. This finding further supports our speculation that hierarchical action control had developed in phases 1 and 2, wherein motor skills for achieving sub-goals had developed in the low-level and memory for sequencing sub-goals had developed in the high-level, and this was facilitated by neuronal noise. 

\begin{table*}[h]
  \caption{Consistency of RNN outputs in representing sub-goals among the last 1,000 episodes in each phase. Data are Mean $\pm$ STD.}
  \label{table:consistency}
  \centering
  \begin{tabular}{lccc}
    \toprule
    \textbf{Network} & Phase 1 & Phase 2 & Phase 3  \\
    \midrule
    {ReMASTER (high level)} &           $\mathbf{0.95\pm0.02}$  & $\mathbf{0.94\pm0.03}$    & $\mathbf{0.95\pm0.02}$ \\
    {ReMASTER (low level) } &           $0.81\pm0.04$           & $0.80\pm0.05$             & $0.80\pm0.05$ \\
    {ReMASTER-single V (high level)} &  $0.88\pm0.06$           & $0.86\pm0.06$             & $0.86\pm0.06$ \\
    {ReMASTER-single V (low level) } &  $0.77\pm0.06$           & $0.80\pm0.04$             & $0.82\pm0.04$ \\
    {ReMASTER-det. (high level)} &      $0.88\pm0.04$           & $0.79\pm0.06$             & $0.85\pm0.04$ \\
    {ReMASTER-det. (low level) } &      $0.75\pm0.05$           & $0.64\pm0.07$             & $0.71\pm0.04$ \\
    {LSTM} &                            $\mathbf{0.85\pm0.20}$  & $0.78\pm0.20$             & $0.54\pm0.29$ \\
    \bottomrule
  \end{tabular}
\end{table*}

\subsection{Consistency in representing sub-goals}
\label{consistency}
To understand the underlying neural mechanisms for ReMASTER's promising performance in relearning phases (Fig.~\ref{fig:Phase2}(c,d)), we analyzed neural data by looking at how consistent the RNN outputs of different RNN architectures could represent sub-goals. 

We measured consistency in representing sub-goals by cosine similarity of temporal profile of $\bm{c}^l$ across the last 1,000 episodes of each phase (see Appendix~\ref{appendix:consistency} for details), for both of the higher level and the lower one (Table~\ref{table:consistency}). It can been seen that higher consistency mostly corresponds to higher success rate for the three models in the consecutive relearning task (Fig.~\ref{fig:Phase2}(b-d)), where ReMASTER agents always showed great consistency in representing sub-goals in the higher level, in contrast to the alternatives, the performance and consistency of which decreased significantly in later phases. We did not compare across different phases because the representation for sub-goals could dramatically change when an agent adapt to a new phase. However, it is rather important that higher flexibility for re-organizing sub-goal representation was shown using ReMASTER agents.

\subsection{Manipulating Agent Behaviors by Clamping High-Level Neural States}

For animals, different brain regions often serve at different levels in action generation. For instance, premotor areas of rodent motor cortex are thought to be important in action choices, while primary motor cortex is considered responsible for details in action execution \citep{morandell2017role}. More interestingly, experimental studies have demonstrated that action primitives of animals can be altered by electrophysiological stimulation or optogenetic inactivation to certain upstream neurons \citep{vu1994identification, morandell2017role}. 

\begin{figure}[h]
    \centering
    \includegraphics[width=0.48\textwidth]{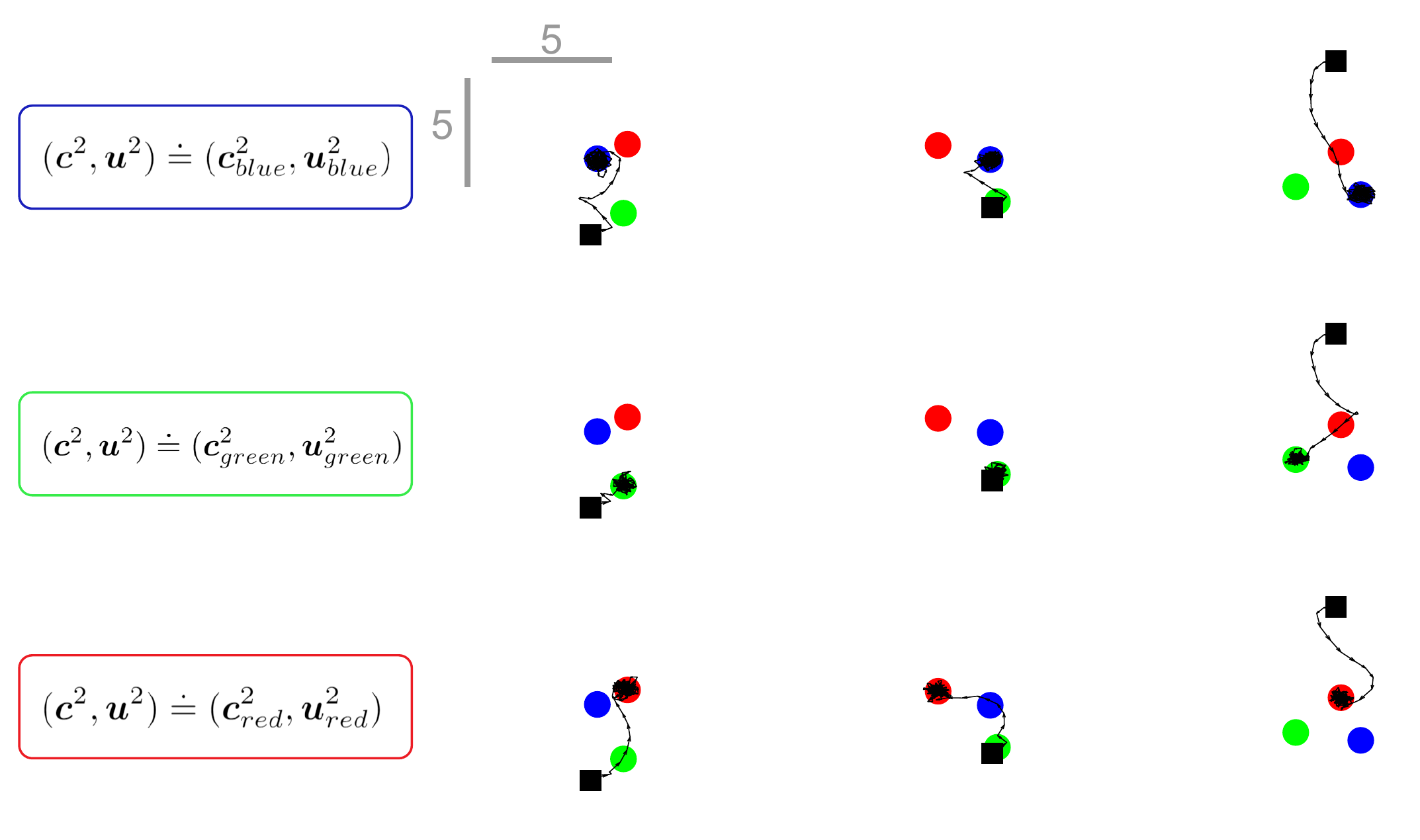}
    \caption{Manipulating agent behaviors by clamping high-level RNN states. All trajectories were from one agent and each row used same high-level RNN states. Black squares and colored circles indicate the agent's initial positions and the target positions, respectively. Each column used the same random seeds for generating initial position and target positions.}
    \label{fig:manipulate}
\end{figure}

Here, we consider analogous experiments on artifacts with ReMASTER agents. We first randomly picked an agent after finishing the consecutive relearning task and then computed the average of $\bm{c}^2$ and $\bm{u}^2$ over the last 500 episodes of phase 3, at the middle step of (i) from initial position to the blue target; (ii) from the blue target to the green one; (iii) from the green target to the red one. By clamping high-level RNN states $(\bm{c}^2$, $\bm{u}^2)$ to those of (i), (ii), or (iii), we could ``manipulate'' a trained agent to consistently follow an action primitive pursuing the corresponding sub-goal (Fig.~\ref{fig:manipulate}). In contrast, fixing low-level RNN states only results in a constant (noisy) action, which is directly determined by $\bm{c}^1$. Therefore, the high-level RNN states act as a label for the action primitives. The continuous property of the RNN states enables representation of an arbitrary number of sub-goals, where in our case we can readily find 3 meaningful action primitives corresponding to the 3 targets.

\label{manipulate}

\subsection{Timescales and Discountings}

\begin{figure*}
    \centering
    \includegraphics[width=0.67\textwidth]{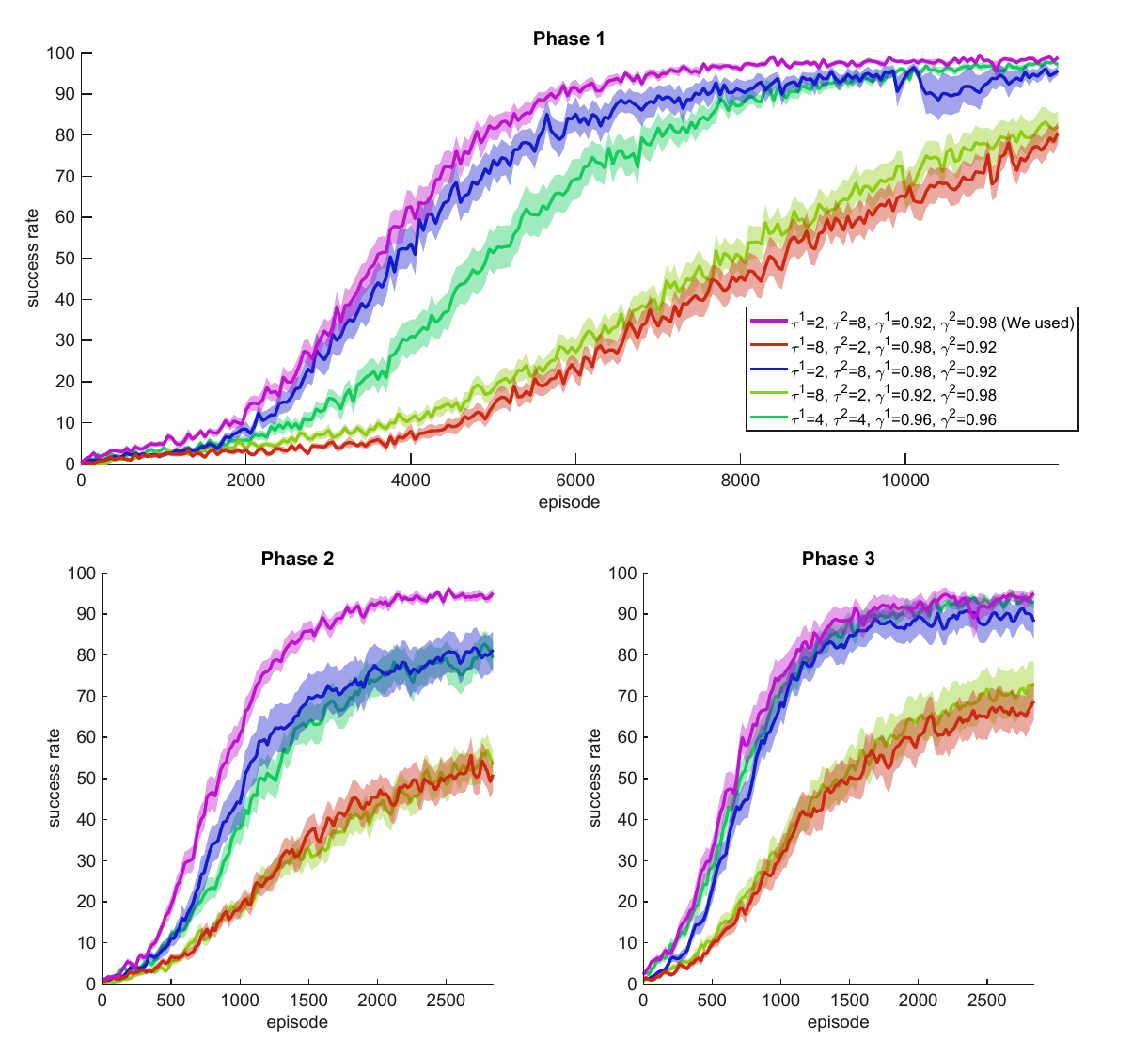}
    \caption{Performance comparison among different settings of $\tau^l$ and $\gamma^l$. Each result was obtained from 10 repeats.}
    \label{fig:taugamma}
\end{figure*}

We have been discussing the role of multiple timescales, indicated by $\tau^l$, the time constant of the $l$th-level RNN, and $\gamma^l$, the discount factor of the $l$th-level value function. In our experiments using ReMASTER, the lower level had smaller $\tau^1$(=2) and $\gamma^1$(=0.92), corresponding to a fast dynamic, whereas the higher level was characterized by a slower timescale ($\tau^2=8, \gamma^2=0.98$). However, a computationally validation of this ``the-higher-the-slower'' setting should also be conducted.

For this purpose, different settings of $\tau^l$ and $\gamma^l$ were examined in the consecutive relearning task. The simulation results (Fig.~\ref{fig:taugamma}) demonstrated a clear advantage of the setting we used, compared to other cases in which ``the-higher-the-slower'' was not followed. Exchange of values of $\gamma^1$ and $\gamma^2$ resulted in significant performance degradation, while alternating values of $\tau^1$ and $\tau^2$ showed even worse performance. Also, it appeared as an unsatisfying choice to set medium values of $\tau$ and $\gamma$ for both layers. The results suggested that a ``the-higher-the-slower'' setting in our model corresponds should be adopted for better performance, which agrees with neurobiological experiments that the higher-level brain regions have longer intrinsic timescales \citep{murray2014hierarchy}. 

\section{Related Work}

Despite much early effort spent on hierarchical RL \citep{sutton1999between, dietterich2000hierarchical} using pre-defined action hierarchies, a number of recent studies have been focused on discovering action primitives\footnote{``Action primitive'' in our paper has a similar meaning as \textit{option} \citep{sutton1999between} in some related literature \citep{bacon2017option,brunskill2014pac, bacon2017option, fox2017multi}} that serve for hierarchical RL. More recent works \citep{brunskill2014pac, bacon2017option, fox2017multi, riemer2018learning} were extensions of the \textit{option} framework \citep{sutton1999between}, which introduced a termination variable to determined start and end of an action primitive. Their works required a pre-defined number of options, whereas our framework can learn to represent an arbitrary number of options by high-level RNN states (Fig.~\ref{fig:manipulate}). Moreover, most of these studies focused on tasks that do not require long-term credit assignment or memorization. In this paper, we consider a different scheme wherein the agent needs to accomplish a series of sub-goals in a particular sequence without observable information indicating the current sub-goal. Such a scheme is common in real life, but has rarely been investigated in RL.

Another track of related studies is skill sharing/reuse among similar tasks in RL. Studies have been conducted considering various schemes, such as meta-RL \citep{santoro2016meta, finn2017model, al2017continuous, finn2018probabilistic, kim2018bayesian, wang2018prefrontal} and lifelong RL \citep{silver2013lifelong, mankowitz2016adaptive, rusu2016progressive, tessler2017deep, abel2018policy}. In particular, some authors proposed ideas shared with our work. Several studies \citep{santoro2016meta, al2017continuous, wang2018prefrontal} employed RNNs for their meta-learning competency, and the others \citep{brunskill2014pac, tessler2017deep} suggested that reuse of action primitives can enhance lifelong learning. However, many of these works considered a multi-task setting where the agent repetitively interacts with a random task sampled from a task set. In our case, the agent first learned to solve one task (phase 1), and the self-developed action hierarchy facilitated relearning in new tasks (phases 2 and 3) with which the agents had never interacted. 

\section{Conclusion and Future Work}

In the current study, we focused on a type of sequential compositional task and comprehensively investigated how they can be solved by autonomously developing sub-goal structure with acquiring necessary action primitives via RL. For this purpose, we proposed a novel RL framework, ReMASTER, which is characterized by two essential features. One is the multiple timescale property both in neural activation dynamics and reward discounting, which is inspired by neuroscientific findings \citep{newell2001time, huys2004multiple, smith2006interacting, murray2014hierarchy, runyan2017distinct}. The other is stochasticity introduced in neural units in all layers, also inspired by the corresponding biological facts \citep{beck2008probabilistic, beck2012not, orban2016neural}.

Simulation results showed that action hierarchy emerged by developing an adequate internal neuronal representation at multiple levels. We presented several pieces of evidences showing that compositionality developed in the network by taking advantage of multiple timescales: abstract action control in terms of sequencing of sub-goals developed in the higher level, while a set of action primitives as skills for detailed sensory-motor control for achieving each sub-goal acquired in the lower level.

Furthermore, compositionality developed in the previous learning enabled efficient relearning in adaptation to changed task goals that involved re-composition of previously learned sub-goals. This re-composition capability was further enhanced with introduction of neuronal noise in addition to motor noise. Such adaptation became possible because development of hierarchical control using multiple levels allowed enough flexibility for re-composition of previously learned control skills.

Since our experiments showed that exchange of timescales between the higher and lower levels resulted in significant performance degradation, it should be worth investigating how an optimal timescale for each level can be determined autonomously during task execution. One possibility is to incorporate LSTMs cascaded in multiple levels \citep{vezhnevets2017feudal} with the expectation that the forgetting gate in LSTMs could provide the means for adaptive timescale modulation.

Finally, ReMASTER is flexible in adopting any gradient-based actor-critic algorithms. Performance can be further improved by employing well-designed model-free algorithms such as \citep{wang2016sample, haarnoja2018soft}. Also, Recent model-based RL methods have addressed promising performance using probabilistic state transition models \citep{deisenroth2011pilco, ha2018recurrent, kaiser2019model}. In this respect, it should be interesting to combine ReMASTER with probabilistic inference of state transitions, using, e.g., a multiple timescale Bayesian RNN \citep{chung2015recurrent, ahmadi2019novel}. Such trials will be attempted in future studies.

\section*{Acknowledgements}

This work was supported by Okinawa Institute of Science and Technology Graduate University funding, and was also partially supported by a Grant-in-Aid for Scientific Research on Innovative Areas: Elucidation of the Mathematical Basis and Neural Mechanisms of Multi-layer Representation Learning 16H06563. We would like to thank Katsunari Shibata, Tadashi Kozuno and Siqing Hou for insightful discussions, and Steven Aird for helping improving the manuscript.

\bibliographystyle{icml2019}
\bibliography{example_paper}

\newpage
\begin{appendices}
\renewcommand{\thesection}{A\arabic{section}}
\renewcommand{\thetable}{A\arabic{table}}
\renewcommand{\thefigure}{A\arabic{figure}}

\setcounter{figure}{0}
\setcounter{equation}{0}
\setcounter{table}{0}

\section{Task settings}

\subsection{Sequential Target-Reaching Task}
\label{appendix:sequential}
\begin{figure}[h]
    \centering
    \includegraphics[width=0.48\textwidth]{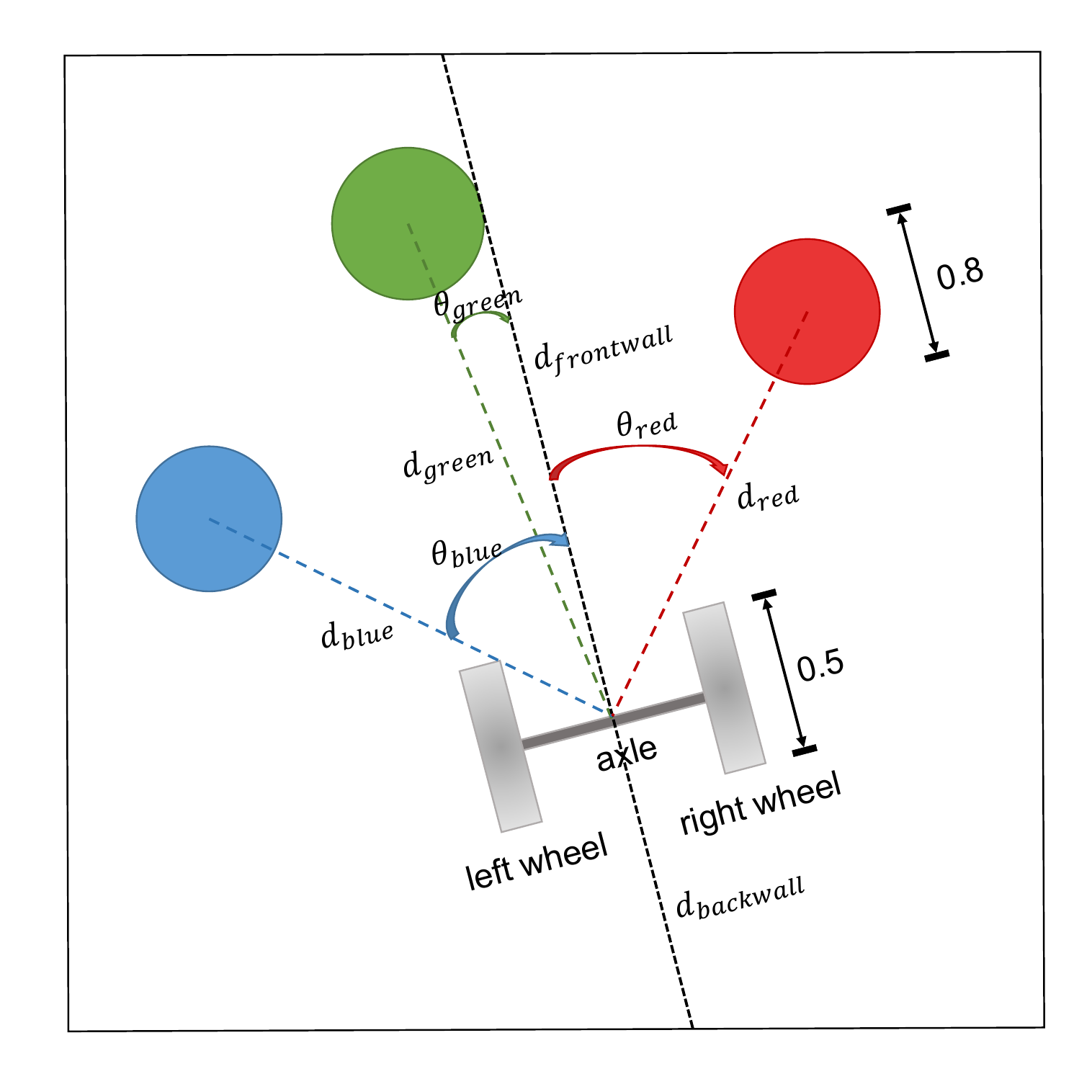}
    \caption{Top view of the sequential target-reaching task. Objects are zoomed out for visual clarity.}
    \label{fig:RGB}
\end{figure}

The setting for a sequential target-reaching task is similar to that in \citep{utsunomiya2008contextual}. A two-wheel robot is required to approach three targets in a sequence, on a $2$-dimensional field, as showed in Fig.~\ref{fig:RGB}. The $2$-D field is a $15 \times 15$ square area, restrained by walls. The robot agent has two wheels of radius $0.25$, connected by an axle. It receives sensory signals to detect distances and angles to the target as well as the walls, as shown in Fig.~\ref{fig:RGB}. At each step, the action is given as the rotations of two wheels, which are continuous in range $[-180^\circ, 180^\circ]$. Length of the axle is $1$, so that the robot can turn $90^\circ$ at most in one step. There are three targets, indicated by red, green and blue, each of which is a circular area of radius $0.4$. At the beginning of each episode, positions of three targets are randomly set inside the center 8 $\times$ 8 area. The distance between two targets are ensured to be larger than $2$. The observation is a $12$-D real number vector: $(e^{-d_{red}/5}$, $e^{-d_{green}/5}$, $e^{-d_{blue}/5}$, $e^{-d_{frontwall}/5}$, $e^{-d_{backwall}/5}$, $r$, $\sin{\theta_{red}}$, $\cos{\theta_{red}}$, $\sin{\theta_{green}}$, $\cos{\theta_{green}}$, $\sin{\theta_{blue}}$, $\cos{\theta_{blue}})$, where $r$ is the immediate reward at current time step, and other quantities are shown in Fig.~\ref{fig:RGB}.

The robot must reach the three targets in the sequence red-green-blue to maximize rewards. The reward function is given as:

If $d_{red}(\tau) > 0.4 \, \forall \, \tau < t$ and  $d_{red}(t) \leq 0.4$, then
\begin{equation}
    r(t) = 0.8/(1+d_{red}(t)).
\end{equation}

If $\exists \, \tau < t$ that $d_{red}(\tau) \leq 0.4$, and  $d_{green}(\tau) > 0.4 \, \forall \, \tau < t$, $d_{green}(t) \leq 0.4$, then
\begin{equation}
    r(t) = 2.0/(1+d_{green}(t)).
\end{equation}

If $\exists \, \tau < t$ that $d_{red}(\tau) \leq 0.4$, and  $\exists \, \tau < t$ that $d_{green}(\tau) \leq 0.4$, and $d_{blue}(t) \leq 0.4$, then
\begin{equation}
    r(t) = 5.0/(1+d_{green}(t)) \quad \mbox{(task done)}.
\end{equation}

If the robot hits the walls, a negative reward $-0.1$ is given. Otherwise the reward is zero.

\subsection{Consecutive Relearning Task}
\label{appendix:relearning}
In the consecutive relearning task, sensory observations for the robot are given consistently across all 3 phases. However, the only difference is that the reward function in each phase appears different. In phase 2, the robot needs to reach three targets according to the sequence green-blue-red, and in phase 3 the required sequence is blue-green-red. Phase 1 had 12000 episodes, while phases 2 and 3 had 3000.

\section{Implementation Details}

\label{appendix:detail}
These experiments were run on CentOS Linux machines using 12-core 2.50 GHz Intel Xeon E5-2680v3 processors. Codes were written with Python 3.6 using TensorFlow library. We used an MTSRNN with 2 levels for ReMASTER in all experiments, where $\tau^1 = 2$ and $\tau^2 = 8$. The discount factors are $\gamma^1 = 0.92, \gamma^2 = 0.98$, computed from $\gamma^l = 1 - 0.16/\tau^l$. There are 100 neurons in the lower level and 50 in the higher level. We directly used the observations as input to the low level RNN. We applied truncated BPTT of length 25 for the sequential target-reaching task, as well as the consecutive relearning task.




Two separate RMSProp optimizers with decay 0.99 were used to minimize the losses of actor and critic respectively, where learning rates were 0.0003 for $\set{L_v}$ and 0.0001 for $\set{L_a}$. We used a replay buffer of maximum size 500,000 and performed experience replay every 2 steps, using a mini-batch containing 16 dispersed sequences with length 25, randomly sampled from the buffer (See Appendix~\ref{appendix:sampling}).

Most of the hyper-parameters were obtained by random search, and we summarized the hyper-parameters used in this study in Table~\ref{table:remaster-hp}. However, a different choice of hyper-parameters does not significantly change our main conclusions in Section~\ref{chap:phase23result} (Appendix.~\ref{appendix:rp}). 

\begin{table*}
  \caption{Hyper-parameters we used in the sequential goal reaching task and the consecutive relearning task for ReMASTER.}
  \label{table:remaster-hp}
  \centering
  \begin{tabular}{lll}
    \toprule
    Hyper-parameter    & Description     & Value \\
    \midrule
    $\gamma^1$  & Low-level discount factor                     & 0.92    \\
    $\gamma^2$  & High-level discount factor                    & 0.98 \\
    $\tau^1$    & Low-level RNN timescale                       & 2 \\
    $\tau^2$    & High-level RNN timescale                      & 8 \\
    $N^1$ & Number of neurons in the lower level                  & 100 \\
    $N^2$ & Number of neurons in the higher level                  & 50 \\
    buffer\_Size & Number of steps recorded in the memory        & 500,000 \\
    $\sigma_0$      &    Initial scale of neuronal noise    										& 0.2  \\
    train\_interval        & The number of steps to run per update 					                    & 2      \\
    learning\_rate\_v  & Learning rate of using RMSProp for critic     			& $0.0003$  \\
    learning\_rate\_a  & Learning rate of using RMSProp for actor      			& $0.0001$  \\
    $\alpha$           & Decay of the RMSProp optimizers      			& $0.99$  \\
    batch\_size      & Number of training sequences per update.        							& 16  \\    
    $L$ & Sequence length for truncated BPTT in training & 25 \\  
    \bottomrule
  \end{tabular}
\end{table*}

\subsection{Replay Buffer for Dispersed Replay}
\label{appendix:sampling}
To enable experience replay, we stored state transitions ($s_t, s_{t+1}, a_t, r_t, done_t$) and RNN states ($\bm{c}_t, \bm{\mu}_t$) in a replay buffer.  We also recorded behavior policy $\pi_t$ in it to compute the importance sampling ratio. We did not separate episodes in the replay buffer. Instead, we consecutively recorded every step, and padded $L-1$ steps at which gradients were not calculated, when an episode terminated. Then, we could randomly sample $n$ sequences of length-$l$ as a minibatch for truncated BPTT, with sampling bias.

\subsection{Initial RNN states for Experience Replay}
\label{appendix:initial}
Different from feedforward neural networks, RNNs for off-policy RL have some practical problems. One major problem is how to decide initial states when training a sequence sampled from the replay buffer. When dealing with finite horizon (episodic) RL tasks, applicable approaches can be summarized as:

\begin{itemize}
    \item \textbf{Recording the RNN states at each step.} RNN states can be treated as hidden observations used in training, which need to be recorded in the replay buffer. Despite the simplicity of this approach, it is unclear what algorithmic issues will be introduced by difference between old internal representations and new ones.
    \item \textbf{Using an entire episode as a sequence.} This was used, e.g., in \citep{mnih2016asynchronous}, providing zero initial states for all the episodes. However, this implementation is computationally inefficient when the length of some episodes is large.
    \item \textbf{Using random sequences with zero initial states.} Sample sequences are randomly sampled from the entire memory, given all-zero initial states. This approach was used in \citep{hausknecht2015deep} for experiments in Atari Games. Unfortunately, this implementation prevents learning long-term dependence because of the mismatch of initial states, as argued in \citep{kapturowski2018recurrent}.
    \item \textbf{Replaying the sequences.} Starting with zero initial states at each episode, RNN states for off-policy updates can be obtained by unrolling new RNNs on old trajectories. A modified version of this approach is offered in \citep{kapturowski2018recurrent}, where the authors assume that computing the forward dynamic of RNNs can help them find better RNN states from zero or recorded RNN states, starting e.g., 20 steps before the start of a sampled sequence. Although \citep{kapturowski2018recurrent} demonstrated remarkable performance on many RL tasks using this approach, their assumption has not been systematically discussed.

\end{itemize}

For simplicity, we employ the first approach. Our experimental results show that it is practical. However, how to choose better initial states still remains a challenge for RL with experience replay using RNNs.

\subsection{Noise Scale}

For continuous sensory-motor tasks, the range of state space can be very large. To enhance efficiency of motor exploration, we used motor noise generated by the \textit{Ornstern-Uhlenbeck process} \citep{uhlenbeck1930theory} (OU-process), like that used in \citep{lillicrap2015continuous}. The OU-process generates temporally auto-correlated noise; thus, the exploration range can be increased with the ``inertia" of the noise. However, it is not necessary to apply temporally correlated noise to hidden states of the MTSRNN, since recurrent connections in an RNN autonomously generate it.

\begin{figure*}
    \centering
    \includegraphics[width=1.0\textwidth]{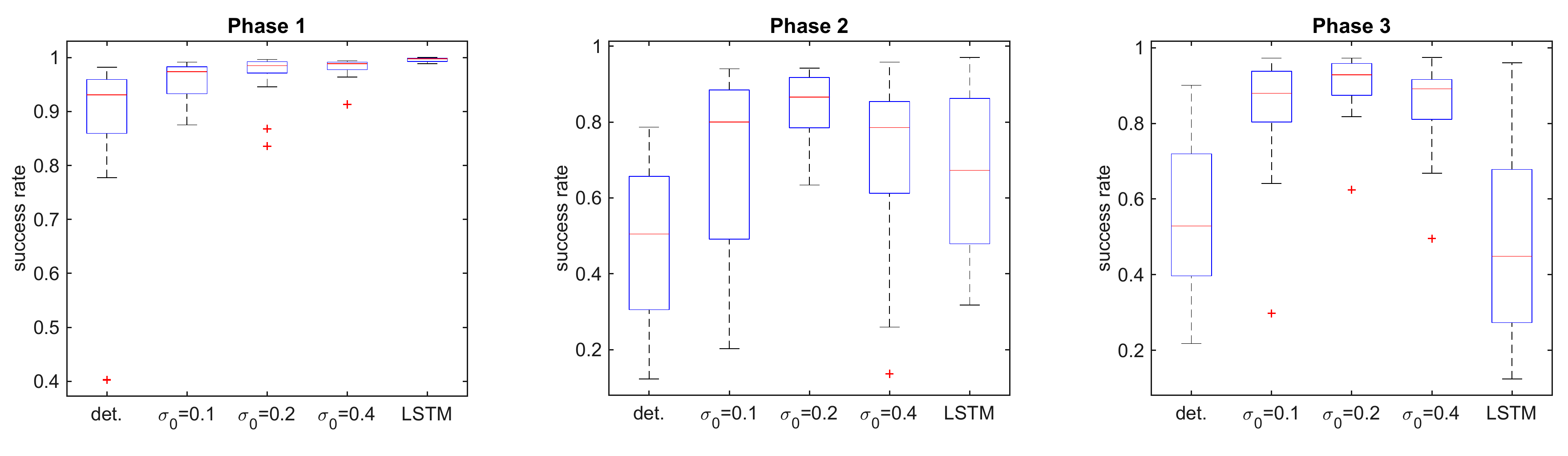}
    \caption{Final performances of ReMASTER for different scales of neuronal noise obtained from last 1,000 episodes of each phase.}
    \label{fig:gg}
\end{figure*}

For exploration in all the tasks, we applied auto-correlated Gaussian noises to the robot's actions, which were generated by independent OU-processes. Each action noise $x_t$ can be computed by
\begin{equation}
x_t = -\theta_{a} x_{t-1} + e \sqrt{2\theta_{a}} \epsilon_t ,
\end{equation}
where $\theta_{a} = 0.3$ for all of our experiments, and $\epsilon_t$ is a unit of Gaussian white noise. $e$ indicates the scale of action noise, which was annealed exponentially w.r.t. episodes, with a minimum value of $0.1$:
\begin{equation}
    e = 180^\circ \times \left[0.75 \times \exp(- \frac{1}{3000} \times \mbox{episode}) + 0.1 \right] .
\end{equation}
Meanwhile, neuronal stochasticity is given by Gaussian white noise with scale
\begin{equation}
    \sigma = \sigma_0 \exp(- \frac{1}{3000} \times \mbox{episode}).
\end{equation}
We performed experiments to determine the proper value of $\sigma_0$, and found that $\sigma_0=0.2$ gave rise to better performance (Fig.~\ref{fig:gg}). Thus we used $\sigma_0=0.2$.

For the consecutive relearning task, at the beginning of phases 2 and 3, we cleared the memory buffer and reset the noise scale, annealed as
\begin{equation}
    e = 180^\circ \times \left[0.75 \times \exp(- \frac{1}{750} \times \mbox{episode}) + 0.1 \right],
\end{equation}
and
\begin{equation}
    \sigma = \sigma_0 \exp(- \frac{1}{750} \times \mbox{episode}).
\end{equation}

\section{Data Analysis}





\subsection{Consistency in Representing Sub-Goals}
\label{appendix:consistency}
This section describes how we computed consistency in representing sub-goals (Section~\ref{consistency}) by cosine similarity of the RNN outputs $\bm{c}$ across different episodes. Because there are usually different numbers of time steps in each episode, we first normalized the number of time steps to 30 for all successful episodes. The normalized time step $t_{norm}=1$ when the agent starts an episode, and $t_{norm}=10, 20$ and $30$ when the agent reaches the first, second, and third target, respectively. Then $\bm{c}^l_{t_{norm}}$ can be obtained w.r.t. the normalized time steps by linear interpolation. Therefore, if the higher-level RNN outputs can consistently represent the correct sub-goals, their temporal profiles w.r.t. normalized time steps $\bm{c}_{t_{norm}}^{2}$ should be similar among different episodes $e$ after convergence. 

Then cosine similarity of $\bm{c}^l_{t_{norm}}$ was then computed for each agent by
\begin{align}
    \mbox{Consistency}^l = \mbox{Mean}_{e_i \neq e_j, t_{norm}, k} \left[ \mbox{CosSim}\left( {c}_{e_i,k,t_{norm}}^{l} , {c}_{e_j,k,t_{norm}}^{l} \right)\right],
\end{align}
where $\bm{c}_{e,k,t_{norm}}^{l}$ indicates the temporal profile (using the normalized time step $t_{norm}$) of the RNN output of the $k$th neuron of the $l$th level in the $e$th episode. $e_i$ and $e_j$ are successful episodes in the last 1,000 episodes of each phase.

\section{Supplementary Results}

\subsection{Effect of Hyperparameters}
\label{appendix:rp}
\begin{figure*}[ht]
    \centering
    \includegraphics[width=1.0\textwidth]{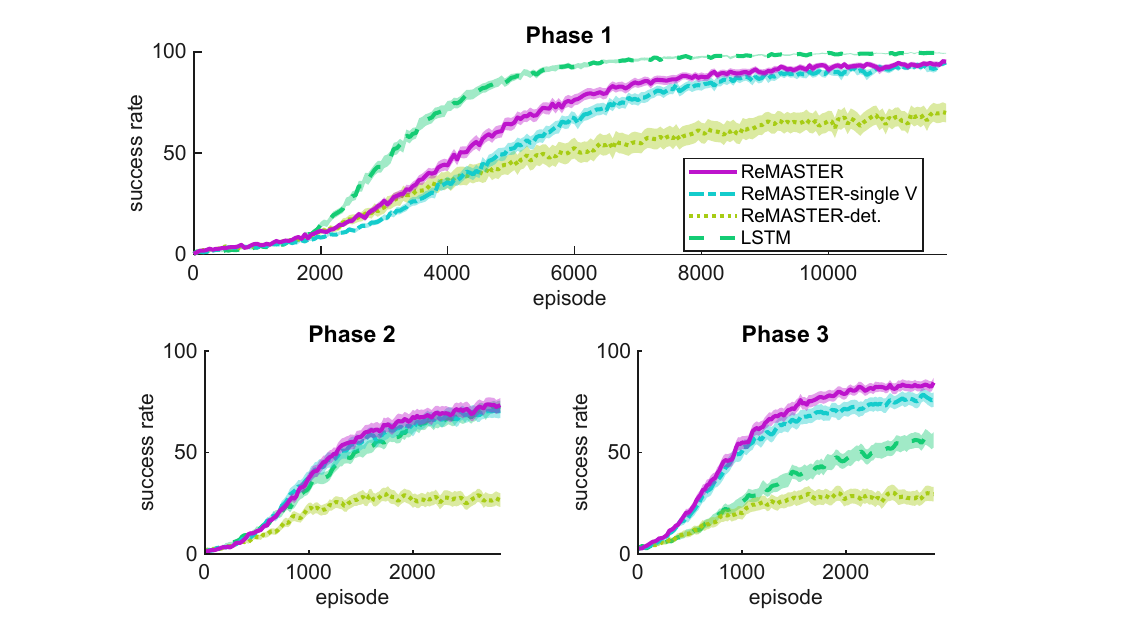}
    \caption{Performance in the consecutive relearning task, using a range of hyperparameters.}
    \label{fig:rp}
\end{figure*}

Many RL algorithms suffer from a proper choice of hyperparameters (such as learning rate, number of neurons in the network) in terms of a satisfying performance. It is also important for us to make sure that the our main results are robust to hyperparameters. For this purpose, we did a random search for hyperparameters (Fig.~\ref{fig:rp}). More specifically, the sequence length for BPTT was sampled log-uniformly in [10, 40]. The learning rate for the actor and for the critic was sampled log-uniformly in [0.00015, 0.0006] and [0.00005, 0.0002], respectively. For the MTSRNN, the number of neurons was log-uniformly in [25, 100] in the lower level, and [50, 200] in the higher-level. The number of LSTM cells was in [40, 160], also log-uniformly sampled. 

As shown in Fig.~\ref{fig:rp}, although the overall performance was a little worse than that using tuned hyperparameters, our conclusions in Section.~\ref{chap:phase23result} did not vary. The LSTM alternative performed better in phase 1, but became worse in the latter phases. Also, ReMASTER always outperformed ReMASTER-det. and ReMASTER-single V. 

\subsection{Performance Gain with Inherited weights}

\label{appendix:control}
\begin{figure*}[h]
    \centering
    \includegraphics[width=0.6\textwidth]{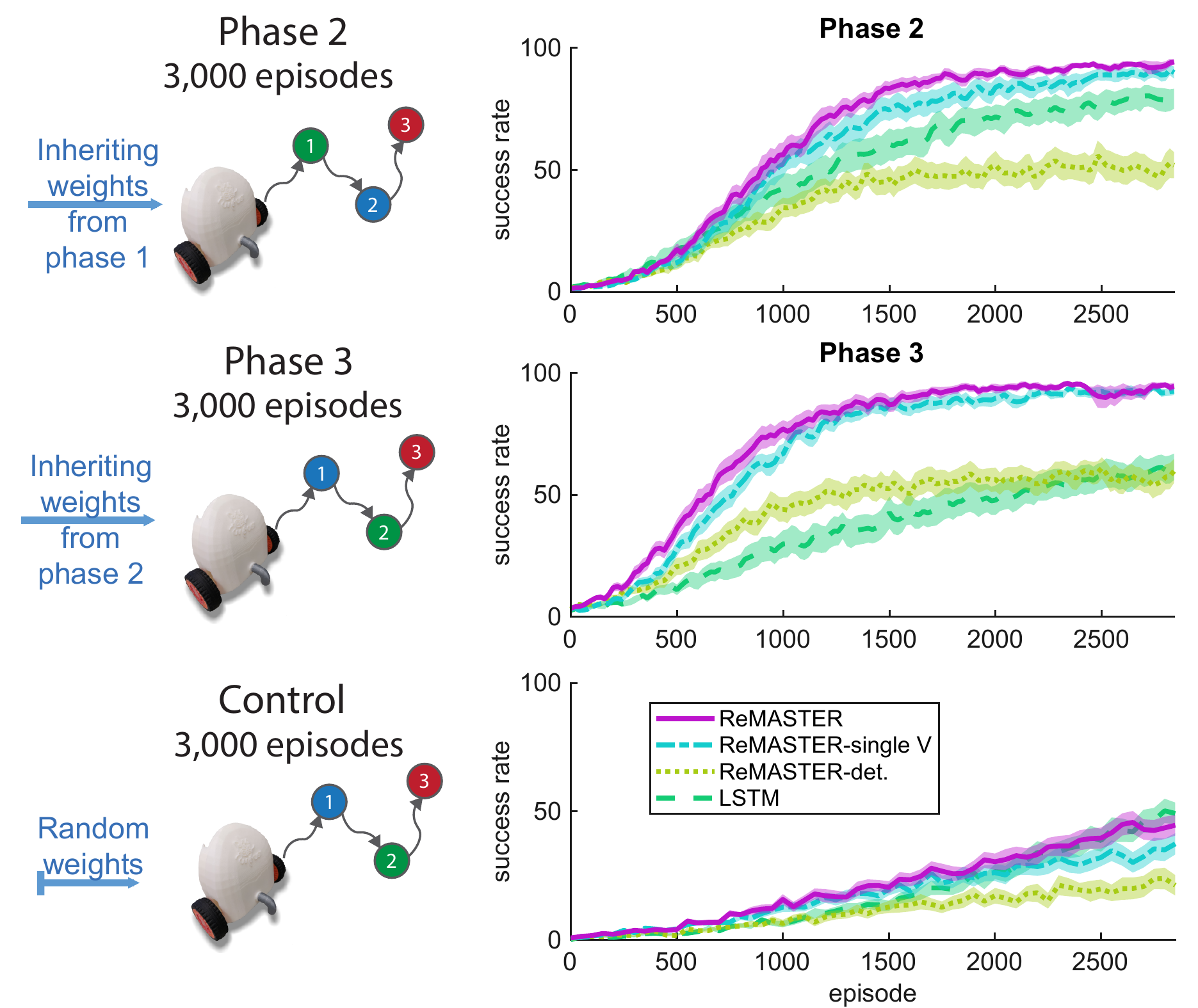}
    \caption{Performance comparison among phases 2, 3 and the control case.}
    \label{fig:control}
\end{figure*}

We prepared a control task that is equal to phase 3 (also equivalent to phase 2 because of symmetry of the three targets) except a random initialization of synaptic weights at the beginning (Fig.~\ref{fig:control}, Bottom). It can be seen that agents with inherited weights largely outperformed agents in the control case that start from scratch, showing meta-learning competency of RNNs \citep{wang2018prefrontal}.

\subsection{Neuronal Noise Ablation Study}
\label{appendix:ablation}

\begin{figure*}[h]
    \centering
    \includegraphics[width=0.75\textwidth]{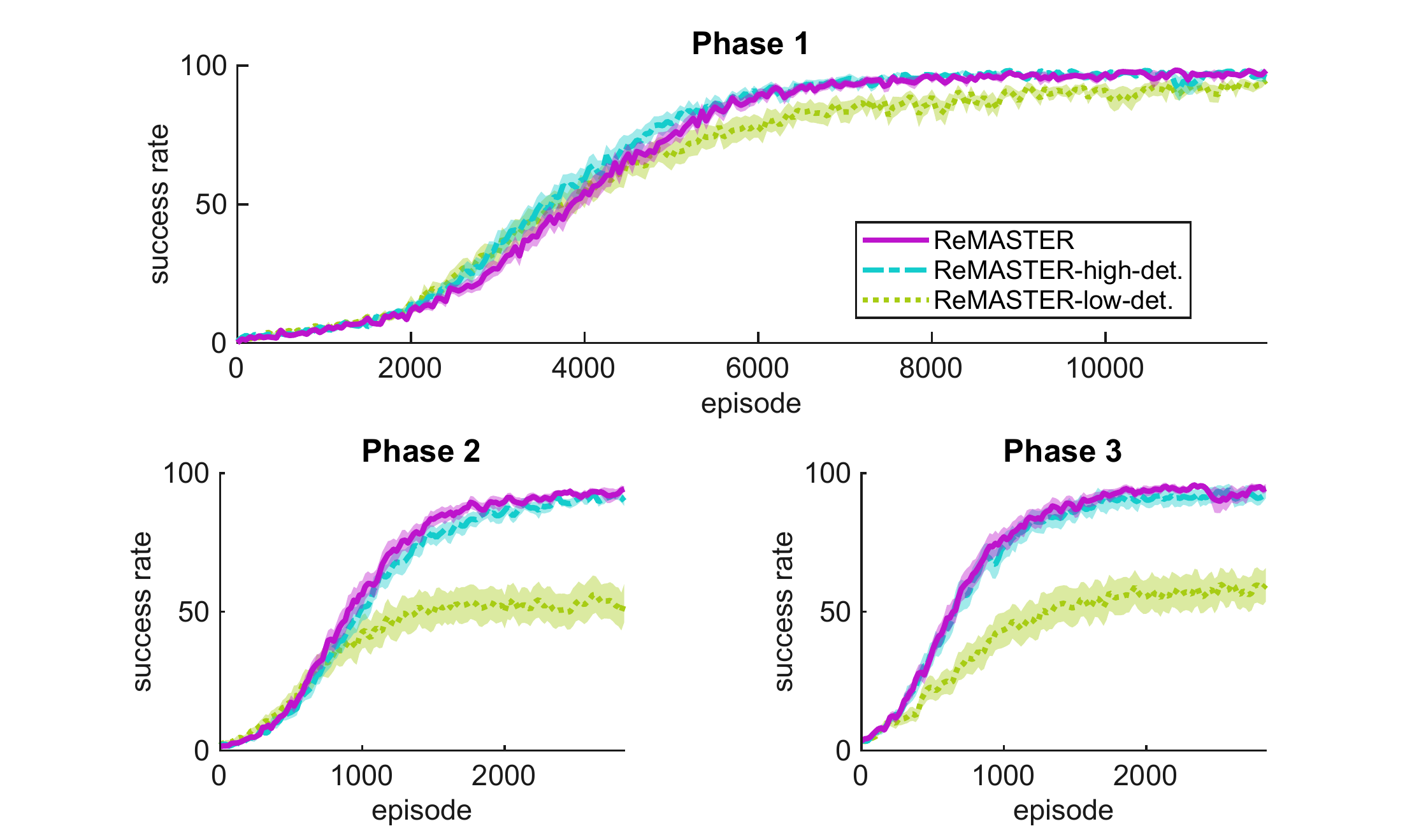}
    \caption{Success rate in all 3 phases. ReMASTER is compared to ReMASTER-high-det. and ReMASTER-low-det., in which the higher-level or the lower-level neurons are deterministic.}
    \label{fig:noise_supplementary}
\end{figure*}

We further conducted experiments to investigate the role of neuronal noise in either the higher level and the lower level. The results (Fig.~\ref{fig:noise_supplementary}) show that, lack of neuronal noise in the higher level lead to slightly worse performance in relearning phases. When the lower-level neuronal followed deterministic dynamics, although it learned slightly faster in phase 1, significant performance degradation was observed in phases 2 and 3. Also, lack of the higher-level stochasticity lead to slightly worse performance in all 3 phases.

\newpage

\begin{figure*}
    \centering
    \includegraphics[width=1.0\textwidth]{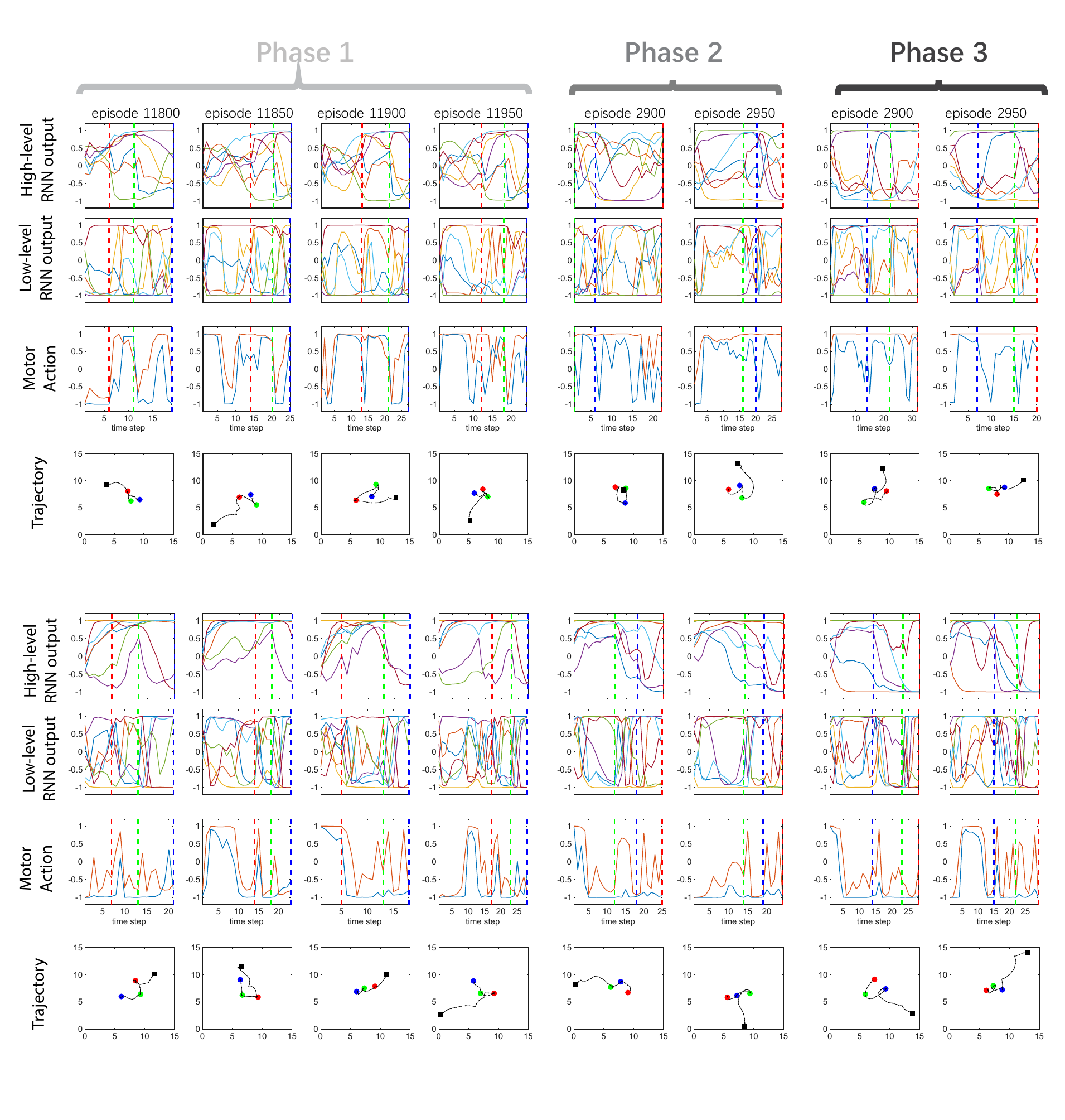}
    \caption{Example episodes showing the behavior of two well-trained ReMASTER agents, in all 3 phases. Plotted in the same way as Fig.~\ref{fig:Phase1_ana}(a) in the main paper. For clarity, the first 7 neurons are plotted for both levels, with different colors indicating different neurons.}
    \label{fig:moreplot}
\end{figure*}

\end{appendices}

\end{document}